\documentclass[11pt,a4paper]{article}
\usepackage[hyperref]{acl2021}
\usepackage{times}
\usepackage{latexsym}

\usepackage{microtype}
\usepackage{graphicx}
\usepackage{multirow}
\usepackage{array}
\usepackage{amsfonts}

\aclfinalcopy 

\title{UPB at SemEval-2021 Task 7: Adversarial Multi-Task Learning for Detecting and Rating Humor and Offense}

\author{Răzvan-Alexandru Smădu, Dumitru-Clementin Cercel, Mihai Dascalu\\
   University Politehnica of Bucharest, Faculty of Automatic Control and Computers\\
  {\tt razvan.smadu@stud.acs.upb.ro} \\
  \tt \{dumitru.cercel, mihai.dascalu\}@upb.ro}
\date{}

\begin{document}
\maketitle
\begin{abstract}
Detecting humor is a challenging task since words might share multiple valences and, depending on the context, the same words can be even used in offensive expressions. Neural network architectures based on Transformer obtain state-of-the-art results on several Natural Language Processing tasks, especially text classification. Adversarial learning, combined with other techniques such as multi-task learning, aids neural models learn the intrinsic properties of data. In this work, we describe our adversarial multi-task network, AMTL-Humor, used to detect and rate humor and offensive texts from Task 7 at SemEval-2021. Each branch from the model is focused on solving a related task, and consists of a BiLSTM layer followed by Capsule layers, on top of BERTweet used for generating contextualized embeddings. Our best model consists of an ensemble of all tested configurations, and achieves a 95.66\% F1-score and 94.70\% accuracy for Task 1a, while obtaining RMSE scores of 0.6200 and 0.5318 for Tasks 1b and 2, respectively.
\end{abstract}


\section{Introduction}
\label{section:introduction}

Sentiment analysis studies expressed opinions and feelings, affective states, and subjective information introduced while conveying ideas to others. In recent years, social media has encountered a major change in ways of interaction and in the freedom of expressing moods to others. For example, individuals frequently use emojis, or even abbreviations, to emphasize their feelings. Therefore, new machine learning systems are developed to predict sentiment valences; however, this task is not trivial, as the detection of sentiments is challenging even for humans. This is mainly generated by the diversity of contexts and differences among people, such as socio-cultural status, age, or gender. For example, certain jokes might be confusing due to past experiences or the current mood while reading the joke. Similarly, language ambiguity can introduce difficulties in understanding the text and in altering the perceived emotions.

Moreover, individuals express a wide range of sentiments, such as happiness, humor, or anger. Words that are used to express such feelings, might share multiple valences and meanings, depending on the context. Therefore, a sequence of words that is, for example, humorous in a certain context, can be considered offensive in another situation. Such examples can rely on bad jokes regarding a person's position in society, ethnicity, or political ideology. Another characteristic that impacts a text's level of humor is how clear the idea is conveyed to the reader.

In this regard, SemEval-2021 Task 7 - HaHackathon: Detecting and Rating Humor and Offense \cite{meaney2021hahackathon} introduced four sub-tasks:  \textit{Task 1a} (classification) includes the class of the text, which can be either humorous or not; \textit{Task 1b} (regression) presents the level of humor as a value between 0 and 5; \textit{Task 1c} (classification) covers the rating of the class of being controversial (i.e., some annotators considered a text to be humorous, while others considered the opposite); \textit{Task 2} (regression) contains the level of offensiveness from the text (i.e., how offensive is the text as an aggregate score agreed by all annotators).

Each example in the dataset contains four labels, one for each task. The labels for Tasks 1b and 1c depend on whether the text contains humor or not. Task 2 is independent of Task 1 and all examples from the dataset have assigned scores corresponding to their level of offensiveness.

In this work, we introduce an adversarial multi-task learning \citep{liu2017adversarial} architecture to detect humor in texts, as well as other related sub-tasks. We focus on Tasks 1a, 1b, and 2, while using the data for all tasks during training. Our model considers BERTweet \citep{nguyen2020bertweet} for contextualized embeddings, followed by Long Short-Term Memory (LSTM) \citep{LSTM} and Capsule layers \citep{sabour2017dynamic} as feature extractors. We observe that adversarial learning increases performance for certain choices of hyper-parameters. Moreover, training the branch for Task 1c on the data for Task 1a improves the performance for Task 1a. In addition, an ensemble method increases the overall performance on all tasks.

The paper is organized as follows. Section \ref{section:related_work} presents related work associated with the provided tasks. Section \ref{section:proposed_method} describes our method, followed by details on the experimental setups and results. Finally,  Section \ref{section:conclusion} provides conclusions and future research paths.


\section{Related Work}
\label{section:related_work}

Several studies \cite{ROBERTS12.201,marasovic2018srl4orl,zaharia2020upb} address the problem of detecting sentiments from texts by using multiple related tasks to further improve the performance of the model. Adversarial multi-task learning was introduced by \citet{liu2017adversarial} and consists of using multiple branches for each individual task (i.e., private branches), and another branch shared among all tasks (i.e., shared branch). A discriminator trained to distinguish between features from multiple tasks is used to separate shared and task-related features in the latent representation from each branch. This is the Adversarial Shared-Private Multi-Task Learning (ASP-MTL) framework that was also employed in other works such as \citep{marasovic2018srl4orl}, where the goal was to label opinions and the associated semantics from texts. Results showed that  ASP-MTL might not achieve better results when compared to classical MTL, and there are other factors that should be taken into consideration, such as dataset split and hyper-parameters.

Other works \cite{zhou2019emotion,spiliopoulou2020eventrelated} used the multi-task technique alongside adversarial learning, but in other configurations than ASP-MTL. For example, \citet{zhou2019emotion} employed a model that has a shared feature extractor, and then it is followed by branches for each task with an attention mechanism for the first layer from each branch. By adding new branches to the model, as well as adversarial learning, the F1-score increased when compared to the baseline model. \citet{spiliopoulou2020eventrelated} adopted domain adaptation to improve the performance of a classifier. Their idea was to add a discriminator after the feature extractor, as a separate branch, that is used to distinguish between different domains. The experiments showed that adversarial learning aided in reducing the bias found in the dataset, thus increasing the model's capability to generalize.

There are multiple Natural Language Processing (NLP) techniques that can be effectively employed to perform text classification in general, as well as sentiment analysis as a specific task \cite{paraschiv2019upb, tanase2020detecting, tanase2020upb, paraschiv2020upb}. For example, neural network methods can be used to detect propaganda in articles by pre-training models on related tasks \cite{vlad2019sentence} or emotions in memes using multimodal multi-task learning \cite{vlad2020upb, vlad2020upbevalita}. Other methods rely on classical machine learning methods (e.g., Support Vector Machine, Naive Bayes, or Random Forest classifiers) and can be successfully used for related tasks, such as fake news detection, obtaining accuracies over 90\% on specific datasets \cite{dumitru2019fake, busioc2020literature}.


\section{Method}
\label{section:proposed_method}

\subsection{Corpus}
The provided dataset for SemEval-2021 Task 7~\citep{meaney2021hahackathon} consists of 10,000 texts (6,179 texts  are considered to have humor, while the rest do not present humor), written in English, with different degrees of humor, and labeled by different categories of people; therefore the labeling is highly subjective. Given the texts with humor, the scores for the Task 1b follow a Gaussian distribution, with the mean of 2.24 (out of 5) and the standard deviation of 0.56. Also, approximately 46.39\% of the texts for Task 1c were labeled as controversial. Furthermore, 42.46\% of the texts were rated as not being offensive at all for the last task, while the 75th percentile is 1.55 (out of 5) for the offensive samples. This means that the data is biased towards not being offensive for the Task 2.

The dataset was already split into train/dev/test sets, such that 8,000 samples were used for training, 1,000 for development, and 1,000 for testing, following similar distributions. Only the entries from training and development sets were provided with annotations during the competition. The annotations for the test set were provided after the evaluation phase ended.

\subsection{Neural Architecture}
\label{section:architecture}
An increasing trend of applying adversarial learning alongside MTL leads to improvements in baseline models \cite{liu2017adversarial, marasovic2018srl4orl, 9207163}. The underlying intuition for this setting is to concurrently use  potential information hidden in the correlation between multiple tasks through MTL, while storing this representation in a shared-private architecture.

\begin{figure}[ht]
\centering
\includegraphics[width=0.48\textwidth]{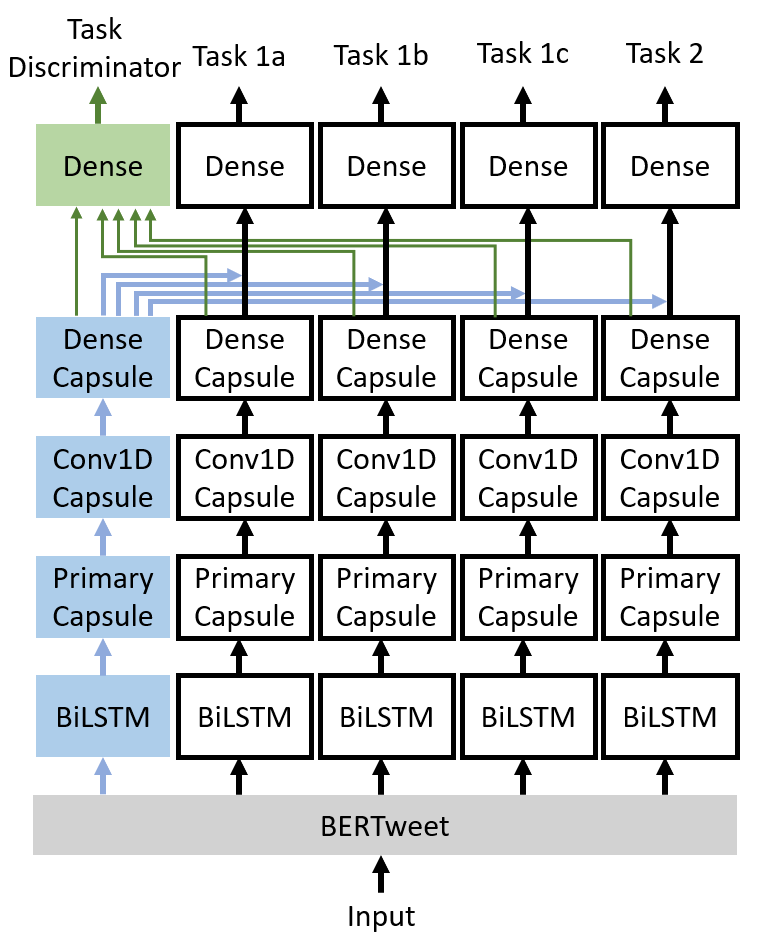}
  \caption{Overview of the AMTL-Humor architecture.}
  \label{fig:AMTL_Humor}
\end{figure}

Inspired by the three previously mentioned works, Figure \ref{fig:AMTL_Humor} shows an overview of our proposed adversarial multi-task learning architecture, namely \textit{AMTL-Humor}. Blue denotes the shared branch among all tasks, whereas  task-specific branches are in white. The discriminative part from the network is represented in green.

We employ a pre-trained Transformer-based model \cite{vaswani2017attention} for generating contextualized embeddings, namely BERTweet, which was trained on 850M tweets written in English. The input is preprocessed in the same way as the input on which the Transformer was trained on, by using the TweetTokenizer from NLTK toolkit \citep{BirdKleinLoper09}. The representation of the input text given by the BERTweet is further fed into five branches. One branch is shared among all other tasks and learns information that is non-task-specific. The output of this branch is concatenated with each private branch, which is then fed to a linear classifier responsible for outputting the final results. A discriminator classifier takes as input the outputs from all branches before the concatenation, having as goal to learn which representation is from what task. In this scenario, labels concerning the branch from which the discriminator took the values are known.

Each branch is composed of five layers, based on a Recurrent Neural Network \cite{cho2014learning} and Capsule layers \citep{8852189, 9207163}. First, the \textit{Bidirectional LSTM (BiLSTM)} layer acts as a feature extractor and encodes the high-level features. It is used to provide a latent representation of the input space that captures the dependencies from both left-to-right and right-to-left. Second, the \textit{Primary Capsule} layer captures the local order of words in the latent space of the BiLSTM output. It encodes the hidden properties, such as positional information for words and the relations between words, from the latent representation into a vector. Third, the \textit{Convolutional 1D Capsule} layer learns the child-parent relationships and predicts the parent Capsules in the next layer by using a dynamic routing algorithm \cite{sabour2017dynamic}. Fourth, the \textit{Dense (Fully Connected) Capsule} layer takes the output of the previous layer, flattens it into a list of Capsules, and then a Capsule is produced to encode the features and their probability by using routing-by-agreement. Fifth, the \textit{Dense (Fully Connected)} layer is the classical Perceptron that takes as input the representation from the last Capsule layer and computes the final output for that branch (which can be for either classification or regression task).

The idea of using Capsule layers was first introduced in Computer Vision  by \citet{sabour2017dynamic} and was studied in various NLP tasks \citep{zhao2018investigating, xiao2018mcapsnet, Wang_2019, 8852189, 9207163, zaharia2020upb}. Capsule networks are based on how human vision works - i.e., Capsules create groups of neurons that are specialized in recognizing certain properties of the input \citep{sabour2017dynamic}.

\subsection{Optimization Problem}
Binary cross-entropy loss is used for the classification problems (i.e.,  humor classification and humor controversy tasks), while the mean squared error loss function is used for  regression tasks (i.e., tasks for predicting humor and offensive scores). Thus, the loss associated with the tasks is computed as follows:
\begin{equation}
	L_{tasks} = \sum_{i=1}^4 \alpha_i L_i
\end{equation}
where $\alpha_i$ represents the weight associated with the task $i$ and $L_i$ is the loss value for the task $i$. 

A problem arises when employing standard MTL, namely, there is a high possibility that part of the shared information may arrive inside the task-related space and vice-versa. This problem can be alleviated by using adversarial learning to enforce the network to make the separation between the private and shared representations. Also, a generalization of the loss function \citep{liu2017adversarial} for training the Generative Adversarial Networks \citep{goodfellow2014generative} is used because we are dealing with multiple tasks, thus multiple classes for the discriminator:
\begin{equation}
	\label{eq:adv_loss_2}
    L_{Adv} = \min_{\theta_S} \left(\lambda \max_{\theta_D} (\sum_{k=1}^5 \sum_{i=1}^{N_k} d_i^k \log [D(E(x^k)))] \right)
\end{equation}

In order to further ensure the separation between the shared and private branches, we add an orthogonality constraint function \citep{bousmalis2016domain} to the objective function during the training, which is defined as:
\begin{equation}
    L_{Diff} = \sum_{k=1}^4 \left|\left| {S^k}^\top H^k \right|\right|_F^2
\end{equation}
where $S^k$ and $H^k$ are the outputs of the from each branch, before the final classification layer, and $\left|\left| \cdot \right| \right|_F$ is the Frobenius norm. 

The total loss for the optimization problem is the sum of all three losses, parametrized by $\lambda$ and $\gamma$, which control the importance for the adversarial and orthogonality losses, respectively:
\begin{equation}
    L_{Total} = L_{Tasks} + \lambda L_{Adv} + \gamma L_{Diff}
\end{equation}

\subsection{Experimental Settings}
\label{section:experimentalsettings}

\subsubsection{Text Preprocessing}
The texts were not cleared; as such  emojis were transformed into text to be further considered as tokens. The TweetTokenizer applies a normalization step to change noise constructs from the text (e.g., URLs) into special tokens. Missing values were replaced with 0 instead of NaN for Tasks 1b and 1c that depend on whether the text was considered to be humorous or not in Task 1a.

\subsubsection{Implementation Details}
A public implementation of the Capsule layer\footnote{\url{https://github.com/naturomics/CapsLayer}} was converted from TensorFlow 1.x to TensorFlow 2.4; this updated version was also publicly\footnote{\url{https://github.com/razvanalex/CapsLayer}}. Also, the gradient reversal layer \citep{ganin2015unsupervised} was used for implementing the adversarial learning between the discriminator and the shared branch, which negates the gradient for the adversarial loss during back-propagation.

The Adam optimizer \cite{kingma2014adam} with a learning rate of 0.001 was considered for training the networks between 10 and 20 epochs, while the batch size was set between 8 and 16. The size of the hidden state for BiLSTMs was set to 128, with a dropout rate of 0.5, as suggested by \citet{JMLR:v15:srivastava14a}. The number of filters in the Primary Capsule layer was set to 8 and the size of the kernel was set to 3x1. Also, the number of filters in the Convolutional 1D Capsule layer was set to 4 and the size of the kernel was set to 3x1. Moreover, the output size was set to 4x1 for all used Capsules, running the routing-by-agreement algorithm for 3 iterations.

During the experiments, different AMTL-Humor configurations were considered by enabling or disabling adversarial learning, the orthogonality constraints, and different parts of the architecture.
Table \ref{tab:training_configs} introduces our configurations, with values lower than 0.5 for hyper-parameters $\lambda$ and $\gamma$, and with conditions concerning adversarial loss being prioritized over the orthogonality loss. 

\begin{table*}[!ht]
	\centering
	\resizebox{0.85\textwidth}{!}{\renewcommand{\arraystretch}{1.1}
\begin{tabular}{m{3.1cm} >{\centering}m{2cm} >{\centering}m{2cm} >{\centering}m{2cm} >{\centering}m{1cm} >{\centering \arraybackslash}m{1cm}}
\hline
\textbf{Model} & \textbf{Adversarial Training} & \textbf{Orthogonality Constraints} & \textbf{Capsule Layers} & \textbf{$\lambda$} & \textbf{$\gamma$} \\
\hline
MTL-Large       & - & - & - & 0 & 0 \\
MTL-Small       & - & - & - & 0 & 0 \\
AMTL-LSTM       & \checkmark & \checkmark & - & 0.1 & 0.01 \\
AMTL-Adv        & \checkmark & - & \checkmark & 0.1 & 0 \\
AMTL-Humor-1    & \checkmark & \checkmark & \checkmark & 0.1 & 0.01 \\
AMTL-Humor-2    & \checkmark & \checkmark & \checkmark & 0.5 & 0.1 \\
AMTL-T1a-Twice$^*$ & \checkmark & \checkmark & \checkmark & 0.05 & 0.01 \\
\hline
\end{tabular}}
\caption{\label{tab:training_configs} The settings for each configuration. The star (*) indicates that the model uses the third branch (i.e., for Task 1c) to solve Task 1a instead. Note that two branches are used for solving Task 1a.}
\end{table*}

\subsubsection{Baseline Models}
Two configurations were used as baseline, namely: a) a larger network (i.e., \textit{MTL-Large}) with 512 hidden states for LSTM cells, 16 primary Capsule filters, and 8 filters for the convolutional Capsule layer with 5x1 filters, and b) a smaller model (i.e., \textit{MTL-Small}) with the previously introduced parameters, sharing the same configurations as the other networks we have tested that use adversarial learning. The optimization step does not involve adversarial learning, nor the orthogonality constraints.

\subsubsection{AMTL-Humor Variants} 
Different models based on the ASP-MTL architecture were tested on the provided dataset. Our base model AMTL-Humor  has two variants (i.e. \textit{AMTL-Humor-1} and \textit{AMTL-Humor-2}), as presented in Table \ref{tab:training_configs}. The only difference between these is the choice for the $\lambda$ and $\gamma$ parameters. Also, \textit{AMTL-Adv} is a model without orthogonality constraints, whereas \textit{AMTL-LSTM} does not contain the Capsule layers in the architecture. Throughout our experiments, we observed that the network always outputs 0 on Task 1c for all inputs; therefore, the model does not manage to learn anything. As such, the \textit{AMTL-T1a-Twice} configuration uses the Task 1a branch twice for both the first and third branches.

\subsubsection{Ensembles}
Ensemble learning was also considered; the modal value (i.e., the most frequent class) is taken  for  classification tasks, while the average of scores is used for regression tasks. \textit{Ensemble-1} combines the results obtained by MTL-Large, AMTL-Humor-1, AMTL-T1a-Twice, and AMTL-Adv respectively, while \textit{Ensemble-2} adds to the previous list the models MTL-Small, AMTL-LSTM, and AMTL-Humor-2.

\subsubsection{Evaluation Metrics}
F1-score and accuracy were used to evaluate the classification tasks, whereas root mean squared error (RMSE) was used to assess performance on the regression tasks from the SemEval-2021 Task 7 competition.

\section{Results}
\label{section:results}

Table \ref{tab:results_dev} presents the results obtained on both development and test sets. On the development set, we observe that increasing the values of the parameters impacts more the regression tasks (i.e., higher RMSE values). Using only adversarial learning, without orthogonality constraints seems to add a small improvement in our case. Removing the adversarial learning has a very small negative impact on Task 1a, and a decrease of RMSE for the regression tasks. Removing only the Capsule layers seems to generate an improvement for this development set. The AMTL-T1a-Twice configuration achieves the highest F1-score due to the symmetry of our architecture. The increase in performance is notable, more than 2\% when compared to AMTL-Humor-1.

\begin{table*}[!ht]
	\centering
	\resizebox{\textwidth}{!}{\renewcommand{\arraystretch}{1.1}
    \begin{tabular}{lcccccccc}
    \hline
    \multirow{3}{*}{\textbf{Model}}  & \multicolumn{4}{c}{\textbf{Development Set}} & \multicolumn{4}{c}{\textbf{Test Set}} \\
                                     & \multicolumn{2}{c}{\textbf{Task 1a}} & \textbf{Task 1b} &  \textbf{Task 2} & \multicolumn{2}{c}{\textbf{Task 1a}} & \textbf{Task 1b} &  \textbf{Task 2} \\
                                     & F1 (\%) & Acc. (\%) & RMSE & RMSE & F1 (\%) & Acc. (\%) & RMSE & RMSE\\
    \hline
    MTL-Large$^*$       & 94.80 & \textbf{93.40} & 0.6434 & 0.7390 & 95.05 & 93.90 & 0.6359 & 0.5891 \\
    MTL-Small           & 93.08 & 90.90 & 0.6747 & 0.6649 & 95.07 & 93.80 & 0.6699 & 0.5796 \\
    AMTL-LSTM           & 94.12 & 92.50 & 0.6939 & 0.6738 & 95.39 & 94.30 & 0.6569 & 0.5552 \\
    AMTL-Adv$^*$        & 93.52 & 92.10 & 0.6979 & 0.7053 & 93.61 & 92.40 & 0.6882 & 0.5631 \\
    AMTL-Humor-1$^*$    & 93.27 & 91.40 & 0.7294 & 0.7294 & 94.93 & 93.80 & 0.7116 & 0.5616 \\
    AMTL-Humor-2        & 93.75 & 92.10 & 0.7426 & 0.6759 & 95.32 & 94.30 & 0.7151 & 0.5680 \\
    AMTL-T1a-Twice  & \textbf{95.88} & 92.09 & 0.6977 & 0.6881 & \textbf{96.29} & 92.85 & 0.6774 & 0.5772 \\
    Ensemble-1$^*$      & 94.05 & 92.50 & \textbf{0.6361} & 0.6656 & 95.66 & 94.70 & 0.6200 & 0.5318 \\
    Ensemble-2          & 94.54 & 93.10 & 0.6383 & \textbf{0.6497} & 95.82 & \textbf{94.90} & \textbf{0.6164} & \textbf{0.5270} \\
    \hline
    \end{tabular}}
    \caption{\label{tab:results_dev} The results obtained on both development set (left) and test set (right). The best scores are marked in bold. The star ($^*$) indicates the models we used for the submissions during the evaluation phase.}
\end{table*}

We observe that our baseline models achieve on the test set similar scores for Task 1a, whereas differences exist for the other two tasks, with no clear better configuration. The larger network tends to learn better on Task 1b, whereas RMSE is worse on Task 2 when compared to MTL-Small. All models that use our adversarial multi-task framework show a small improvement over the baselines on Task 2; nevertheless, almost all models perform worse on Task 1b, especially our system (i.e., AMTL-Humor-1) that registers the highest RMSE scores. All adversarial models achieve or the Task 1a similar results, or even better when compared to the baseline models, and this was our main focus during the training.

The AMTL-T1a-Twice model manages to obtain the highest F1-score on one of the two branches for Task 1a. However, accuracy and F1 scores were low when making predictions for Task 1.c (46.34\% and 61.18\%, respectively). These low performance values may be indicative of bias in the dataset towards either 0 or missing values for this task.

Ensemble models complement the inner configurations and provide more stable predictions. Ensemble-1 was the model submitted during the competition for evaluation. The differences between our Ensemble-1 configuration and the best models reported for each task were the following: 2.88\% less F1-score and 3.5\% less accuracy for Task 1a, 0.124 higher RMSE for Task 1b, and 0.119 higher RMSE for Task 2. During the post-evaluation phase, we assessed the Ensemble-2 configuration that performs better than Ensemble-1 on all tasks by a small margin (i.e., less than 1\%).

\subsection{Visualizations}

t-SNE visualizations \citep{tsne} were considered to better grasp the adequacy of our model. t-SNE is an algorithm used to visualize high-dimensional data into a two-dimensional representation by minimizing the Kullback-Leibler divergence between the joint probability distributions for both lower-dimensional and higher-dimensional data from the input. Since the output from BERTweet has 59,904 dimensions (i.e., 78 tokens in a sequence, each having 768 features given by the hidden units), we first reduced the representation to 100 dimension by applying a Principal Component Analysis \cite{jolliffe2016principal}. This way, t-SNE runs faster, without losing too much information when considering the principal components. 

\begin{figure*}[ht]
\centering
\includegraphics[width=1\textwidth]{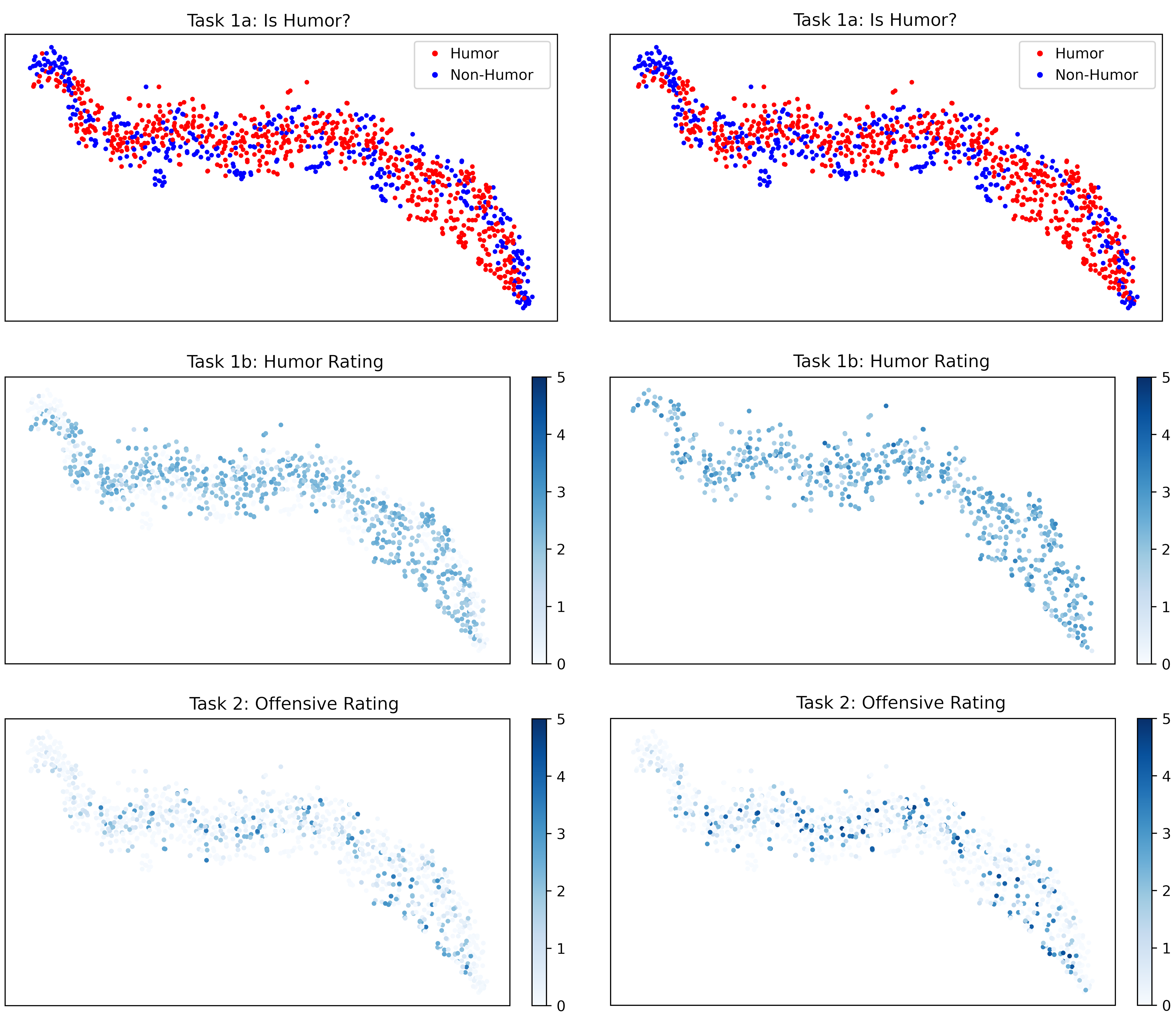}
  \caption{t-SNE projection for the development set data on the output of the BERTweet layer. On left are the plots for predicted outputs, whereas on right are the corresponding plots for ground-truth. Best viewed in color.}
  \label{fig:TSNE_dev}
\end{figure*}

The results using the Ensemble-2 configuration are shown in Figure \ref{fig:TSNE_dev} for Tasks 1a, 1b, and 2. Each point represents the embedding of the input sample, while the colors represent the classes or the values for each sample. Based on the ground-truth visualizations (left side), we can observe that the points are not linearly separable. Our best configuration was capable to learn the inner representation of the input data. For the offensive rating task, the examples with higher scores are in the middle of the manifold. We can also observe for the offensive rating task that the majority of classes are biased towards lower rather than higher values.

\section{Conclusions and Future Work}
\label{section:conclusion}
In this work, we present AMTL-Humor, an adversarial multi-task learning method to deal with the problem of detecting and rating humor and offense for SemEval-2021 Task 7. More specifically, our model is inspired by the ASP-MTL framework and considers BiLSTM and Capsule layers as feature extractors on top of the BERT layer that provides contextualized embeddings. 

Two ensemble models were created using variations of our AMTL-Humor architecture. These configurations were trained either on different settings (such as adversarial learning) or by modifying the structure of the branches. We observed that adversarial learning might perform better than other architectures with similar structures, while considering specific configurations and tasks. Another improvement was observed when the model was trained on two tasks using the same branch, namely the AMTL-T1a-Twice model. This approach of duplicating branches may be used in other context as it adds redundancy to the network and supports generalization. Finally, the best results were obtained by using an ensemble over all trained models.  

In the future, we aim to study how performance can be improved, especially on Task 1c, by adapting the hierarchical multi-task learning technique  \cite{sogaard2016deep} to the AMTL-Humor architecture.


\bibliographystyle{acl_natbib}
\bibliography{anthology,acl2021}

\end{document}